\documentclass[conference]{IEEEtran}
\IEEEoverridecommandlockouts
\PassOptionsToPackage{bookmarks={false}}{hyperref}
\usepackage{cite}
\usepackage{amsmath,amssymb,amsfonts}
\usepackage{algorithmic}
\usepackage{graphicx}
\usepackage{textcomp}
\usepackage{xcolor}
\usepackage {hyperref}
\usepackage{fancyhdr}
\fancypagestyle{alim}{\fancyhf{}\fancyfoot[L]{978-1-7281-8629-0/20/\$31.00 ©2020 IEEE}\lhead{2020 6th Iranian Conference on Signal Processing and Intelligent Systems (ICSPIS)}}
\def\BibTeX{{\rm B\kern-.05em{\sc i\kern-.025em b}\kern-.08em
    T\kern-.1667em\lower.7ex\hbox{E}\kern-.125emX}}
\begin{document}
\begin{NoHyper}
\author{\IEEEauthorblockN{1\textsuperscript{st} Hamed Hosseini}
\IEEEauthorblockA{\textit{Electrical \& Computer Engineering} \\
\textit{University of Tehran}\\
Tehran, Iran \\
hosseini.hamed@ut.ac.ir}
\and
\IEEEauthorblockN{2\textsuperscript{nd} Mehdi Tale Masouleh}
\IEEEauthorblockA{\textit{Electrical \& Computer Engineering} \\
\textit{University of Tehran}\\
Tehran, Iran \\
m.t.masouleh@ut.ac.ir}
\and
\IEEEauthorblockN{3\textsuperscript{rd} Ahmad Kalhor}
\IEEEauthorblockA{\textit{Electrical \& Computer Engineering} \\
\textit{University of Tehran}\\
Tehran, Iran \\
akalhor@ut.ac.ir}
}

\title{\textbf{Improving the Successful Robotic Grasp Detection Using Convolutional Neural Networks}\\
}
\maketitle
\thispagestyle{alim}

\begin{abstract}
Robotic grasp should be carried out in a real-time manner by proper accuracy. Perception is the first and significant step in this procedure. This paper proposes an improved pip line model trying to detect grasp as a rectangle representation for different seen or unseen objects. It helps the robot to start control procedures from nearer to the proper part of the object. The main idea consists in the pre-processing, output normalization, and data augmentation to improve accuracy by 4.3 percent without making the system slow. Also, a comparison has been conducted over different pre-trained models like AlexNet, ResNet, Vgg19, which are the most famous feature extractors for image processing in object detection. Although AlexNet has less complexity than other ones, it outperformed them, which helps the real-time property. 
\end{abstract}

\section{Introduction}\label{introduction}
One of the essential needs of general-purpose robots is interacting and manipulating the objects in their surrounding; this is one of the routines that human beings do without any latency. When they impose an object, they instantly know how to grasp it; however, it is so sophisticated for the even state of the art algorithms to predict the proper grasp completely real-time. General-purpose robots are employed in different production lines; so, during robotic manipulation and grasping, the robot should generate proper grasp suggestions for an unlimited number of objects in different environments, which might be posed to the robot that has never been exposed.

Several researchers have worked on robotic grasping during the last decades \cite{b1, b2, b3, b4}. This problem is challenging related to perception, planning, and control \cite{b4, b5, b6, b7}. The robot should process the image captured by the robot's camera and suggest a proper location and orientation of the end-effector to grasp objects for the desired task. Then the trajectory to reach the goal is planed by the planning module. Finally, the control part should create proper motor commands to the actuators to follow this desired trajectory. In some techniques called robust function detection \cite{b15} and structured grasp detection\cite{b14}, which this paper is based on, the planning and control are separate modules. In contrast, there are some other grasping techniques based on learning visuomotor control policy which perform these two tasks simultaneously\cite{b27}.

The model-based approaches for solving the grasping problem have been discussed for several years \cite{b30}. They were firstly known as hard coding, specializing in a specific task with fixed objects by expert knowledge. Later some other model-based approaches appeared with modeling all the environment, object, and contact. This modeling was standing over object geometry shape, material, mass distribution, environment shape, task space constraints, and contact properties like friction and force. These methods were useful during so many years, but they had some significant limits. They could not model every complicated object, environment, and contact. Also, they could not handle new unseen objects in new shapes since they were not modeled before. Following the advent made on deep learning algorithms, recently grasping techniques have been improved for unseen objects \cite{b13, b14, b15, b16}. In this paper, scene perception using deep learning algorithms improved.

Perception---understanding the scene---is a complicated issue conducted by the sensory information of the robot, like vision, force, distance, and etc. where the most available and valuable sensory data is vision. This paper focuses on the 2D RGB images and the extended one, which includes depth as a new channel called RGB-D image, in which the depth information is retrieved from some Kinect cameras. Imagine that the robot faces a new image which consists of multiple or a single object which they could be in each part of the image. Here there are some main challenges in scene understanding which researchers confront previously in other tasks. Segmentation\cite{b8, b9} is the first one that has been used extensively in the literature for several tasks, such as, medical image processing \cite{b10}, which can help to grasp by segmenting camera image into different parts and objects. Localization \cite{b11}  by the aim of finding the exact location of something in the images was used massively in many tasks like autonomous driving \cite{b23}; now, its application in robotic grasp has been widely used in order to help the robot find the location of the object going to grasp. Detection \cite{b12} and recognition newly applied to the grasp detection problem. Previous works on detection focused on especially some applications like face detection[23] and gesture detection \cite{b22}, but recently using object detection for grasping is favor.
 
In this paper, grasp detection is formulated as a regression problem by representing the appropriate grasp as a 5-dimensional representation as shown in Fig. \ref{gr_rep_fig}. It is simpler than the 7-dimensional representation \cite{b7} and could be mapped to that 7-dimensional in order to execute robotic grasp successfully \cite{b14}. So any input image should be mapped to a 5-dimensional grasp representation, which guarantees a successful grasp. This representation is available in some data sets. One of the main RGB-D images data sets is the Cornell Grasp Detection Data Set shown in Fig. \ref{fig:cornell_dataset}, which is consists of 1035 images of 280 different objects, with several images taken of each object by different orientations or poses.
 
\begin{figure}
\centerline{\includegraphics[width=8cm,height=10cm,keepaspectratio]{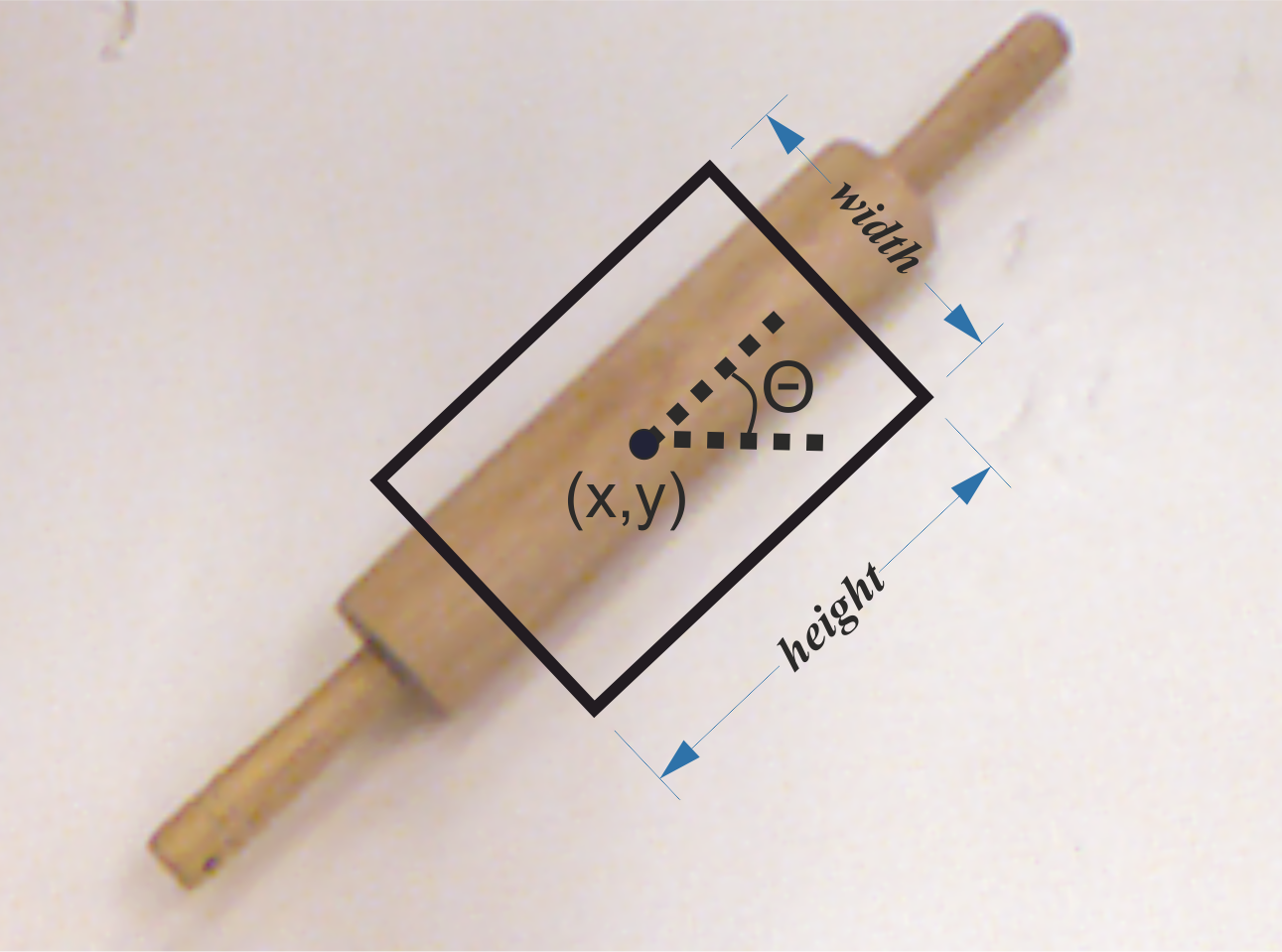}}
\caption{Five-dimensional grasp representation, which consists of location,
orientation and gripper's plates size.}
\label{gr_rep_fig}
\end{figure}

Grasping problem has been appeared in many different tasks and solved by different solutions. The result comparison is essential. Some benchmarks and indexes like ACRV Picking Benchmark (APB) \cite{b18} , Yale-CMU-Berkeley (YCB) \cite{b19} and Jaccard index were performed. The last index is more suitable to the task-free grasping problems and is used in related papers, so the paper results have been compared among this index.

Real-time processing is vital in grasp detection because of the importance of the response to every image. Any delay could make the grasping fail. The processing time of the grasp detection is related to the computer's processing unit capacity and the complexity of the algorithm used. Previous algorithms that used a sliding window were slow and took at least 13.5 seconds per frame with an accuracy of 75 percent\cite{b20}. Since the sliding window algorithm was slow, global view methods which process all the image immediately instead, make the procedure more real-time \cite{b26}. So, in this paper, input image feed into out pip-line architecture is a global view that has been augmented and pre-processed by a different technique. The pipeline architecture consists of the pre-trained and regressor, trying to regress the input image to the 5-dimensional of proper grasp, which is normalized. 

The main contribution of this paper consists in improving the grasp detection accuracy based on the Jaccard index measurement. Moreover, in order to achieve the latter result, some tuning in pre-processing, data augmentation, and output normalization are applied. Finally, the performance of the proposed algorithm, in term of processing time and accuracy, were put into contrast over different pre-trained models, such as AlexNet, VGG19Net, and ResNet.

The remainder of the paper has the following structure.
In Section \ref{problem_statement}, the whole problem is defined. Section \ref{aproach} is devoted to represent the architecture, data set, pre-processing, output normalization and data augmentation.
In Section \ref{experiments}, the measurement and the results can be seen.
Finally, the paper concludes with some hint and ongoing works. 

\begin{figure}
    \centering
    \includegraphics[width=8cm,height=10cm,keepaspectratio]{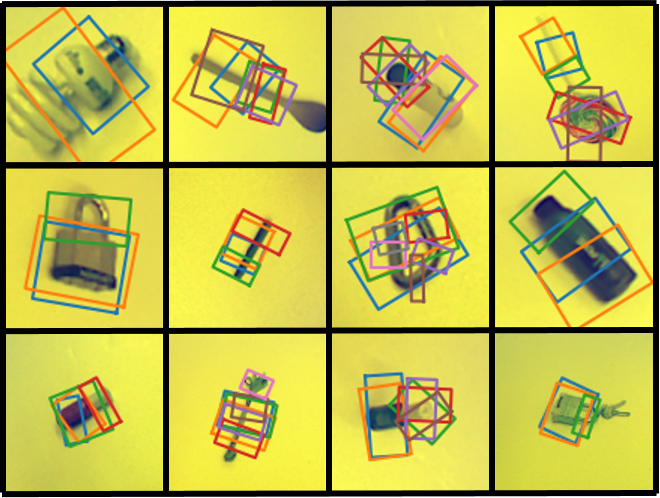}
    \caption{Cornell Grasp Data Set: some of the Cornell grasp image data set with their multiple ground truth labels are shown. The blue channel is replaced with depth information.}
    \label{fig:cornell_dataset}
\end{figure}
\section{robotic grasp problem statement}\label{problem_statement}
The robot's intelligent unit needs to understand the environment and suggest an appropriate grasping offer to the control unit. This paper is concentrated on the perception stage of the robot task. For this goal, a grasp representation should be assigned to any image exposed to the robot. The rectangular representation shown in Fig.  \ref{gr_rep_fig} is a common one that has been frequently used in literature which can be mathematically formulated as follows:

\begin{equation}
gr = \{x, y, \theta, w, h\}\label{gr_rep}
\end{equation}
In the above, $(x, y)$ is the center position of the grasp rectangle, $\theta$ is the orientation of the rectangle corresponding to the $x$-axis. $w$, $h$ stands for width and height of the rectangle respectively. The height of the rectangle depends on the gripper plate size and the width of the rectangle should be smaller than the gripper maximum opening size.

This problem can be regarded as a regression one, mapping the RGB-D image matrix to a grasp vector which is defined as Eq. \eqref{gr_rep}. This is similar to the detection problem excepted to three main differences: first, bounding box orientation estimation is necessary here while it is not essential in detection problem; secondly, this model could find a part of an object as good grasp prediction while it is not possible in detection one. Finally, it is not essential for grasp detection problem to detect the object's name, it is only aim to understand which part of the object is a right candidate to be grasped.

\section{grasp detection approach}\label{aproach}
Deep convolutional neural networks have recently been used in many detection and regression problems. A pipeline model is performed, which gets an RGB-D input and delivers the proper grasp rectangle as output. This structure, which is shown in Fig. \ref{fig:structure} consists of two sequential processing units, which each part has an independent responsibility. The feature extractor is due to extract brief and meaningful properties of an image that can be suitable for finding a proper grasp candidate. As the next step, the regressor part is designed to map these extracted features to a successful grasp suggestion.  

\subsection{Grasp Detection Architecture}
\begin{figure}
    \centering
    \includegraphics[width=8cm,height=10cm,keepaspectratio]{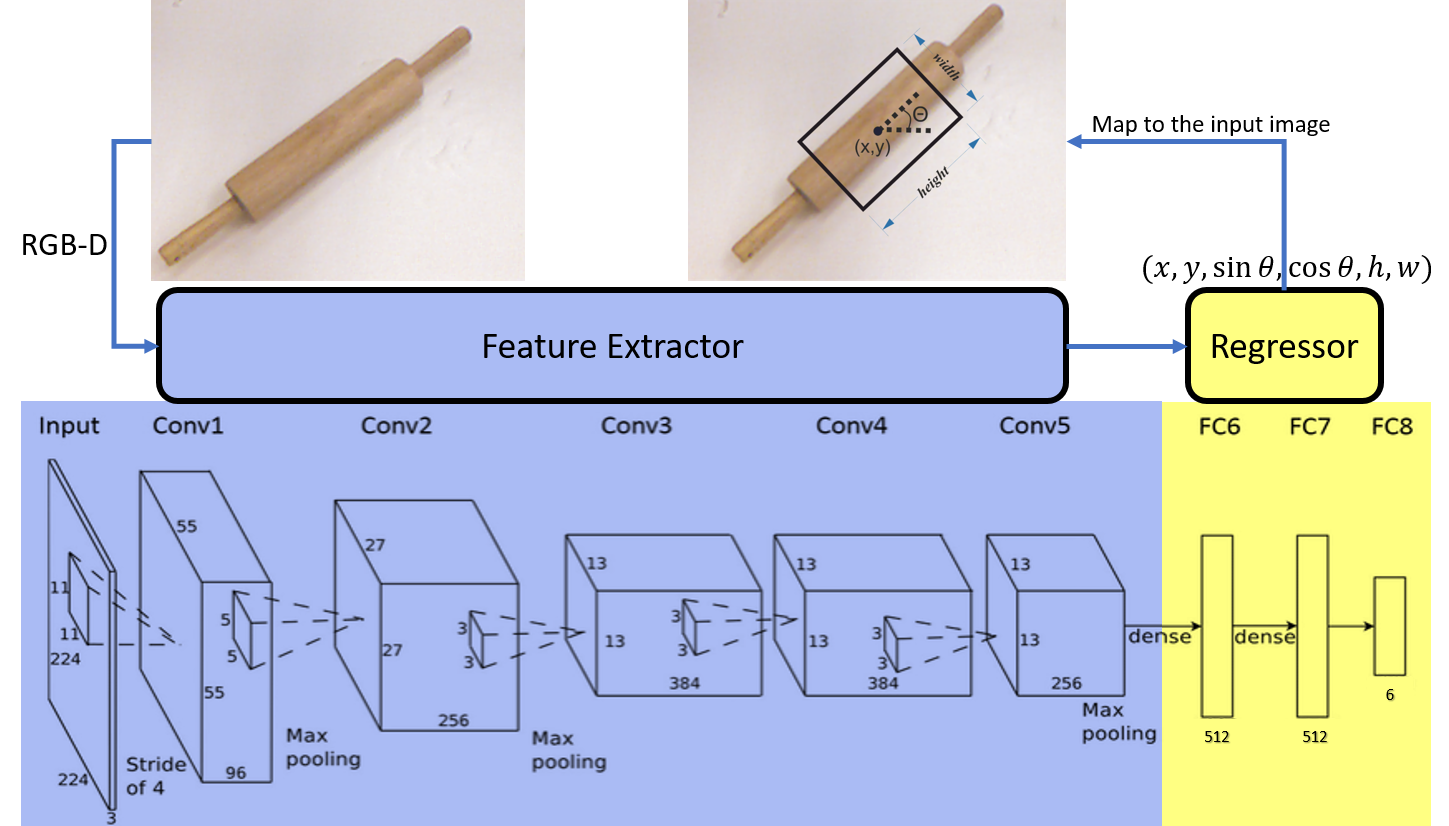}
    \caption{Outline of the pipe-line model: RGB-D as input and grasp representation as output. Each block is extended by the same color.}
    \label{fig:structure}
\end{figure}
The pipe-line model depicted in Fig. \ref{fig:structure} consists in a convolutional network as a first step to get proper features of the RGB-D image and the second step is to regress these features to desire rectangle. According to the input complexity, which is consists of 3 color and one depth channel information, Deep convolutional layers is used in the feature extractor part, but since feature extractor block has reduced the information complexity, the regressor is handled with a more simple neural net. The output of this model, $\hat{t}$, has 6 Neurons  which correspond to the following parameters:
\begin{equation}
    \hat{t} = (\hat{x}, \hat{y}, \hat{\sin\theta}, \hat{\cos\theta}, \hat{w}, \hat{h}) \label{that}
\end{equation}
According to the appearance of the grasp location properties ($x, y$) as two terms in grasp representation, the angle parameter is duplicated in two symmetric rotational terms ($\sin\theta$ and $\cos\theta$) in order to make the balance in paying attention to location and orientation of grasp rectangle.

Since the model's training time is too long and the possibility of over fitting in  deep layers,  different pre-trained models are used. Transfer learning---training or fine-tuning the architecture which is followed by some pre-trained models---has been used previously in many other tasks. Applying the foregoing method to the grasp detection problem, has worked well.
The fully connected layers are designed as three dense layers by the \textit{Tanh} activation function because of the image normalization into $[-1, 1]$. The number of neurons is $512 \times 512 \times 6$ that the output neurons will be corresponds to the rectangle properties. In order to avoid over fitting during training, the dropout layer with a probability of 0.5 is considered between dens layers.
\subsection{Cornell Grasp Data Set}\label{AA}
One of the most important data sets which have been extensively used in robotic grasp detection context is the Cornell grasp data set shown in Fig. \ref{fig:cornell_dataset} \cite{b24}. It consists of 885 images of 240 distinct objects labeled by some ground truth rectangle labels. Each image is labeled by a number of grasps rectangles corresponding to some possible ways to grasp the object which is labeled by humans. This data set is suitable for parallel plate grippers. Although the labels includes a wide range of orientations, locations, and scales, it does not mean that every grasp rectangle is covered in this labeling.

Since only one label is needed during training phase for each image according to the architecture, this paper selected one of the labels for image randomly and the training is performed on the latter label. 
\subsection{Data Pre Process}
 The pre-trained models are trained on the three-channel RGB images, while the Cornell Grasp Data consists of the RGB-D images, which have four channels and need to be feed to the network. Since the depth channel has more information than each color individually, it is replaced into the blue channel to train the fully connected layers.
 
The input RGB-D image of the network consists of 4 channels by the color density  between 0 and 255 and depth density according to the camera location. Each channel of an image is min-max normalized by itself as following formula to scale into $[-1, 1]$ as:
\begin{equation}
    RGB{\text -}D_{\textrm{normed}} = (\frac{RGB{\text -}D - \min(RGB{\text -}D)}{\max(RGB{\text -}D) - \min(RGB{\text -}D)} *2) - 1 \label{eq:normalising}
\end{equation}
Also, each image's dimension has been transformed into $224 \times 224$ in order to be compatible with the input of the feature extractor.
\subsection{Output Normalization}
Since the proposed solution for this grasping detection problem is a regression method, approximate consistent between input and output domain would be meaningful. As discussed above, the network's input image normalized into the interval of $[-1, 1]$. Normalizing the target to the nearly same range of the input, surprisingly improved the accuracy with over 30 percent. Therefore the target and prediction is redefined as follows:
\begin{equation}
    \label{eq:y_norm}
    \begin{aligned}
    t_n &= (x_n, y_n, s_n, c_n, w_n, h_n)
    \\
    \hat{t_n} &= (\hat{x_n}, \hat{y_n}, \hat{s_n}, \hat{c_n}, \hat{w_n}, \hat{h_n}) 
    \end{aligned}
\end{equation}
where $(x_n,y_n)$ is the normalized rectangle center, $w_n, h_n$ are normalized width and height of the rectangle and $ s_n, c_n $ are normalized orientation parameter that wholly re-ranged to $[0, 1]$ by the following formula:
\begin{equation}
\label{eq:norm_param}
\begin{aligned}
    s_n &= \frac{(\sin\theta + 1)}{2}\\
    c_n &= \frac{(\cos\theta + 1)}{2}\\
    x_n &= \frac{x}{224}\\
    y_n &= \frac{y}{224}\\
    w_n &= \frac{w}{224}\\
    h_n &= \frac{h}{224}
\end{aligned}
\end{equation}
It should be noted that the raw values of $x, y, w, h$ are in the range of 0 to 224.
\subsection{Data Augmentation}
The available data is not efficient for the training of these many parameters. A data set in the training flow should be used to train the network. Feeding a picture twice in the training procedure is not valuable, but other images that have been derived from the origin image by a little change can make new information during training. This action is referred to augmentation. Two types of techniques are used during image augmentation, namely, rotation and zoom. In the case of rotation all images are augmented by 8 image rotated by one of these  angles: $[0, \frac{\pi}{4}, \frac{\pi}{2}, \frac{3\pi}{4}, \pi, \frac{5\pi}{4}, \frac{3\pi}{2}, \frac{7\pi}{4}]$.
In the case of zoom, all data are zoomed by some ratio to the image center which helps to predict different images by the different distances from the camera, near, far, and too far. This is done by the scaling ratio between [0.5, 1]. Zoom factor 1 means there is not any zoom in the picture, and zoom by 0.5 means getting $\%50$ of the image. Every image is zoomed by 6 different values: $[0.5, 0.6, 0.7, 0.8, 0.9, 1]$

\section{experiment and evaluation on grasp detection}\label{experiments}
\subsection{Training Dens Layers}
The training procedure is performed by 25 epochs. Every training batch consist of 128 randomly selected images and include ground truth labels and the other training parameters is available in Table \ref{tab1}. 
According to the network architecture, every single image from the data set should be corresponded to a single ground truth rectangle label. Since the Cornell grasp data set contains more than one ground truth grasp label for each input image, it needs one to be chosen. Random selection was done in this step in order not to make the grasp prediction biased to any specific ground truth label. Also a single image may be trained with different labels since it may appear in different batches. This helps the model learns all the edges and essential features for any object grasping. 
Since training of deep convolutional neural networks is time consuming, usually, suitable computational devices are needed. The proposed model has been run on GPU(GeForce 1080, 8G RAM). Code has written by PyTorch on CUDA 9.0 and is available on my GitHub page \footnote{https://github.com/hamed-hosseini/ggcnn}
\begin{table}
\caption{Training Parameters}
\begin{center}
\begin{tabular}{|c|c|c|}
\hline
\textbf{Batch Size}&{\textbf{128}}&\textbf{} \\
\cline{1-2} 
\textbf{Batch per Epoch}&{\textbf{100}}&\textbf{Train Phase} \\
\cline{1-2}
\textbf{Epoch}&{\textbf{25}}&\textbf{} \\
\hline
\textbf{Split Ratio}&{\textbf{0.9}}&\textbf{} \\
\cline{1-2}
\textbf{Optimiser}&{\textbf{Adam}}&{\textbf{common}}\\
\cline{1-2}
\textbf{Loss Function}&{\textbf{MSE}}&\textbf{} \\
\hline
\textbf{Validation Batch Size}&{\textbf{10}}&{\textbf{Validation Phase}}\\
\cline{1-2}
\textbf{Validation Batches}&{\textbf{100}}&\textbf{} \\
\hline
\end{tabular}
\label{tab1}
\end{center}
\end{table}

\subsection{Pre-trained Feature Extractor Models}
Different pre-trained models like AlexNet, ResNet, Vgg19 Bached Norm are used as the Features extractor block in Fig. \ref{fig:structure}.
Pre-trained models are trained with $224 \times 224$ images of the ImageNet data set \cite{b25}, which consists of 14,197,122 images with their bounding box labels used for detection. Since images' essential features can be taken from these networks, a pre-trained network is used as the feature extractor. At the training phase, this part is frozen in order not to be changed, and only the fully connected layer is trained and it designed to regress the rectangle for these extracted features.

Another important point should be mentioned is feeding the depth information to the pre-trained models. As these networks learned with three RGB channels, the depth is extra information in this structure. So depth information is replaced to the blue channel after normalizing to the range $[0, 1]$, then the accuracy calculated by the Jaccard index improved over 5-6 percent. This means that the depth image has more information than the blue one in the regression problem.
The benefit of using pre-trained networks is overfitting avoidance and training time reduction.
\subsection{Grasp Index}
There should be an index for evaluating the grasp performance. Two different metrics have been offered for the Cornell grasp data set. The point metric has previously used in the literature on this data set, which indicates the Euclidean distance between the center of predicted and target rectangles. This metric has some drawbacks such as not considering the orientation. So two conditions were used to evaluate the prediction success:
\begin{itemize}
    \item 
    Angle between predicted grasp rectangle and target rectangle should be lower than 30\textdegree.
    \item
    Jaccard index of the predicted and target grasp rectangle should be more than $25\%$.
    
    Jaccard index is defined as follows:
    \begin{equation}
        J(A, B) = \frac{|A \cap B|}{|A \cup B |}
    \end{equation}
    which means that the overlap area of the predicted and ground truth rectangles should be more than $25\%$ of their union area.
\end{itemize}

\begin{figure*}
    \centering
    \includegraphics[width=15cm,height=18cm,keepaspectratio]{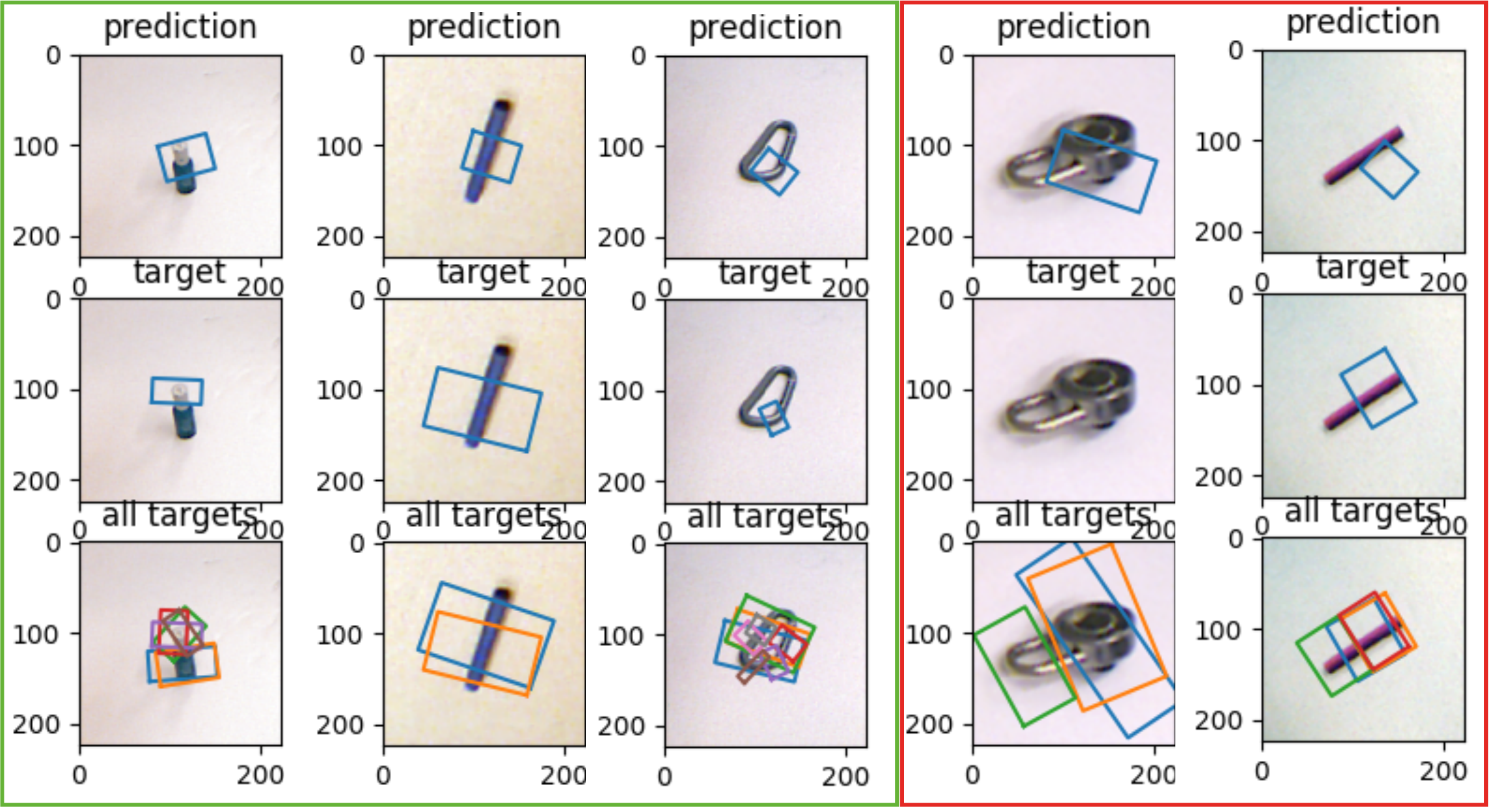}
    \caption{Left green box contains correct predictions, and the red right box contains wrong predictions.
}
    \label{fig:prediction}
\end{figure*}

\subsection{Grasp Detection Results}
\begin{figure}
    \centering
    \includegraphics[width=8cm,height=10cm,keepaspectratio]{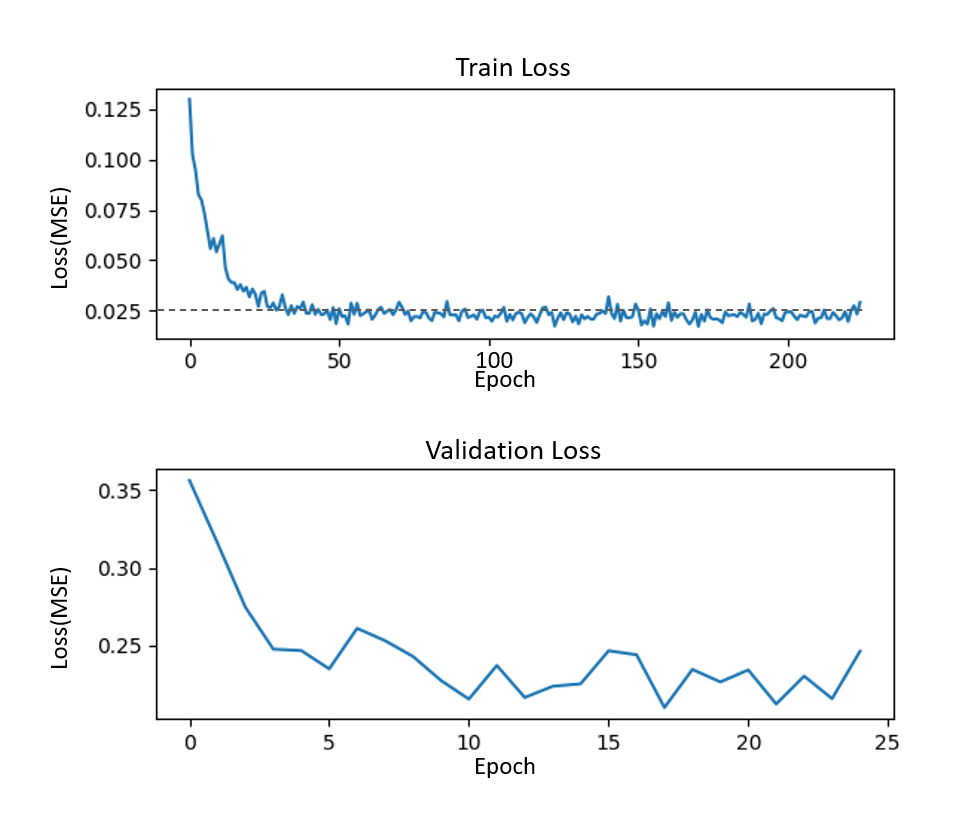}
    \caption{Decreasing train and validation loss function value among increasing training epochs.}
    \label{fig:loss}
\end{figure}
Some proper and lousy grasp predictions are available in Fig. \ref{fig:prediction}. Left green box contains correct and the right red box contains incorrect predictions. As expected, it can be seen that the grasp prediction rectangles are not limited to any specific part of the object and they have freedom to be in any part, some predictions covers whole the object and some related to a small suitable part. This originates from label random selection in the input feeding stage. According to the accuracy defined in the last section, the accuracy of the model with different pre-trained networks has been depicted in Table \ref{tb:Pretraind}. Also this paper detection accuracy could be compared to the baseline algorithms in Table \ref{tb:detection_accuracy}. Decreasing loss function value among increasing the training epochs can be seen in Fig. \ref{fig:loss}. This figure shows that there is not any over fitness, since both training and validation loss are decreasing and also model has been trained over epochs and has reached to a stable train value loss which is over 0.025.\\

\begin{table}
\caption{min, max and mean accuracy of the algorithms}
\begin{center}
\begin{tabular}{|c|c|c|c|}
\hline
\textbf{Pre-trained Network}&{\textbf{min}}&{\textbf{mean}}&{\textbf{max}} \\
\hline 
\textbf{AlexNet} & {\textit{79.6\%}}& {\textit{84.5\%}}&{\textbf{88.7\%}}\\
\hline
\textbf{ResNet50} & {\textit{52.8\%}}& {\textit{54.6\%}}&{\textit{56.9\%}}\\
\hline
\textbf{Vgg19 Batched Norm} & {\textit{78.9\%}}& {\textit{82.4\%}}&{\textit{86.7\%}}\\
\hline
\end{tabular}
\label{tb:Pretraind}
\end{center}
\end{table}

\begin{table}
\caption{detection accuracy comparison.}
\begin{center}
\begin{tabular}{|c|c|}
\hline
\textbf{Approach}&{\textbf{Accuracy}} \\
\hline 
Jiang et al. \cite{b7} & {\textit{60.5\%}}\\
Lenz et al. \cite{b14} & {\textit{73.9\%}}\\
Redmon et al. \cite{b26} & {\textit{84.4\%}}\\
Kumra et al. \cite{b28} & {\textit{84.76\%}}\\
Wang et al. \cite{b29} & {\textit{85.3\%}}\\
\hline
\textbf{Improved method} & {\textit{88.7 \%}}\\
\hline
\end{tabular}
\label{tb:detection_accuracy}
\end{center}
\end{table}

\section{Conclusion}\label{conclusion}
In this paper, an improved real-time grasp detection system with the help of different pre-processing and output normalization methods over possible pre-trained neural networks was performed. It was demonstrated that the depth channel information is more valuable than each RGB channels. Depth information could be used in the pre-trained network which was trained by the RGB images. The latter was achieved by replacing the depth information to any channel. Data augmentation like zooming and rotating the input images in some discretized values can feed new information to the network and help the regressor learn better. Also, output normalization improved accuracy and helped over fitness prevention. While the input and output range was so apart from each other, the model could not be trained accurately, but as soon as output normalization occurred, training and validation loss decreased simultaneously and accuracy increased. It was shown that between available pre-trained networks, the AlexNet performed better by the measure of accuracy and response time. This grasp detection is the first step of grasping procedure, which can be applied to famous arms like universal and Baxter and also famous grippers like Robotiq.
In the future works, the implementation of this algorithm on the ROS and Gazebo platform is aimed, and combining the input sensory image with the force sensors would be valuable. In addition, since it was inferred that location prediction is more accurate than the angle, these two parts can be separated and tune individually which can be regarded as an ongoing work. 

\vspace{12pt}
\end{NoHyper}
\end{document}